\documentclass{article}

\usepackage{natbib}

\usepackage[final]{nips_2017}

\usepackage[utf8]{inputenc} 
\usepackage[T1]{fontenc}    
\usepackage{hyperref}       
\usepackage{url}            
\usepackage{booktabs}       
\usepackage{graphicx} 
\usepackage{amsfonts}       
\usepackage{nicefrac}       
\usepackage{microtype}      
\usepackage{caption}
\usepackage{subcaption}

\title{Malaria Likelihood Prediction By Effectively Surveying Households Using Deep Reinforcement Learning}

\author{
  Pranav Rajpurkar \\
  Department of Computer Science\\
  Stanford University\\
  \texttt{pranavsr@cs.stanford.edu} \\
  \And
  Vinaya Polamreddi \\
  Department of Computer Science\\
  Stanford University\\
  \texttt{vinaya.polamreddi@cs.stanford.edu} \\
  \AND
  Anusha Balakrishnan \\
  Department of Computer Science\\
  Stanford University\\
  \texttt{anusha@cs.stanford.edu}
}

\begin{document}

\maketitle

\begin{abstract}
We build a deep reinforcement learning (RL) agent that can predict the likelihood of an individual testing positive for malaria by asking questions about their household. The RL agent learns to determine which survey question to ask next and when to stop to make a prediction about their likelihood of malaria based on their responses hitherto. The agent incurs a small penalty for each question asked, and a large reward/penalty for making the correct/wrong prediction; it thus has to learn to balance the length of the survey with the accuracy of its final predictions. Our RL agent is a Deep Q-network that learns a policy directly from the responses to the questions, with an action defined for each possible survey question and for each possible prediction class. We focus on Kenya, where malaria is a massive health burden, and train the RL agent on a dataset of 6481 households from the Kenya Malaria Indicator Survey 2015. To investigate the importance of having survey questions be adaptive to responses, we compare our RL agent to a supervised learning (SL) baseline that fixes its set of survey questions a priori. We evaluate on prediction accuracy and on the number of survey questions asked on a holdout set and find that the RL agent is able to predict with 80\% accuracy, using only 2.5 questions on average. In addition, the RL agent learns to survey adaptively to responses and is able to match the SL baseline in prediction accuracy while significantly reducing survey length.
\end{abstract}

\begin{figure}[t]
  \centering
  \includegraphics[width=0.4\linewidth]{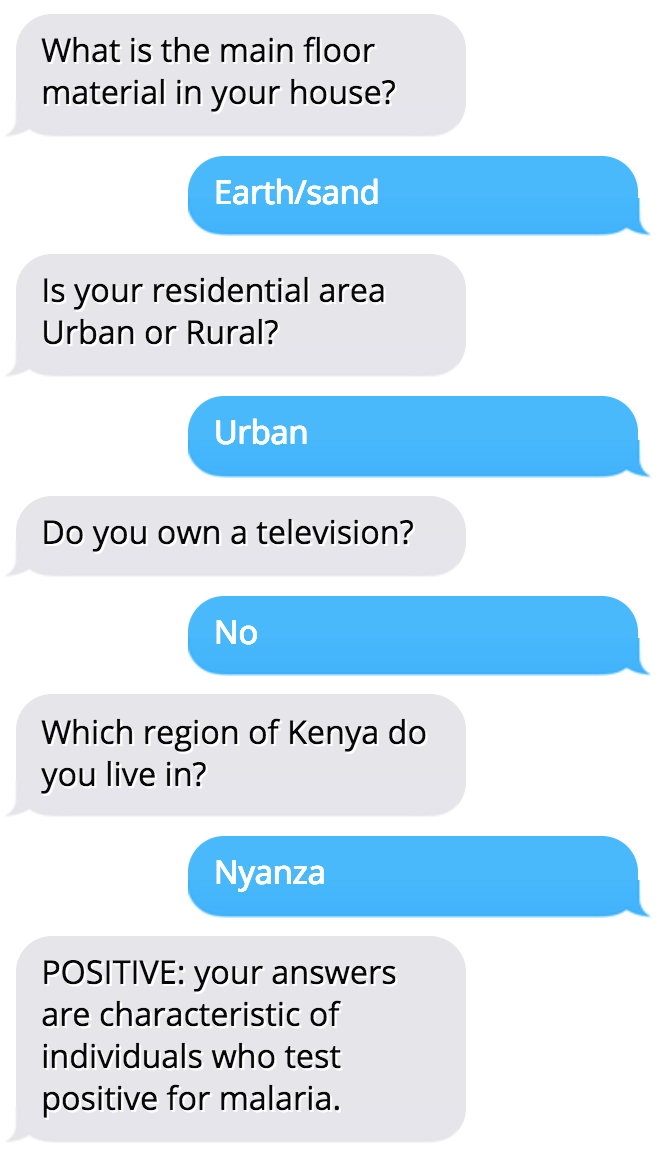}
  \caption{
       The RL agent is able to effectively survey individuals to determine their likelihood of testing positive for malaria by asking them questions about their households. On this example, the agent correctly predicts that the individual being surveyed is malaria positive.
  }
  \label{fig:example}
\end{figure}

\section{Introduction}
We build a short mobile survey questionnaire that informs individuals of their risk of malaria, aiming to improve malaria management at the national and individual level. Making surveys short and available on mobile allows for quicker, cheaper and more timely disease monitoring \citep{nature}. In addition, making people aware of their risk of having malaria promotes health seeking behaviors \citep{who}. We focus on the Malaria Indicator Survey (MIS) in Kenya, which plays a central role in assessing and guiding the public policy for targeted malaria interventions \citep{who}.

We use reinforcement learning (RL) to train an agent that learns to survey individuals to accurately predict their likelihood of testing positive for malaria. The agent is allowed a total of $k$ sequential decisions: at each time it can decide whether to ask the individual another question or make a prediction about the result of their malaria blood smear test based on the survey questions answered. The agent incurs a small negative reward for every question asked, a large negative reward for incorrect predictions, and a large positive reward for correct predictions. The agent must thus learn to balance the cost of asking questions with the benefit of making an accurate prediction.

Our RL agent is a deep Q-network that learns a policy directly from the state of the survey: the responses to the questions posed so far serve as input to a convolutional neural network, which outputs a $Q$ value for every action. We define the action space such that there is an action associated with each possible question that can be asked and with each possible prediction class. To train the RL agent, we use the Kenya MIS 2015, containing detailed surveys with individuals in their households and biomarkers that include the results of their malaria blood smear tests.

We investigate the benefits of a survey that is adaptive to responses by comparing the RL agent against a supervised learning (SL) baseline that fixes the length and the content of the survey beforehand. The SL baseline uses the response to a fixed set of $k$ questions as input to a convolutional neural network, which outputs the probability of testing negative and positive on the malaria test. As $k$ increases the SL model has higher accuracy though it accumulates a higher cost for asking more questions. On a holdout set, we evaluate agents on their prediction accuracies and average survey length, and find that the RL agent learns to predict the result of malaria test with $80\%$ accuracy asking only $2.5$ questions on average, matching the accuracy of the SL baseline, but outperforming it on survey length.

\section{Data}
We use the Kenya Malaria Indicator Survey 2015, a nationally representative survey done to assess the progress of malaria interventions and monitor prevalence across the target population \citep{kmis}. The survey contains data about 6,481 households, including information the use of malaria nets in the household, ownership of household assets such as TV and access to electricity, biomarker test results including a rapid diagnostic test and a microscopic blood smear test that detect the presence of malaria parasites. We focus on the household member recode survey, which has a record for every person in the household.

\subsection*{Processing}

We process the data to keep only categorical variables which have no more than 10 classes, and records for which there is no missing data. Because our focus is on data that can be collected remotely through mobile surveys, we filter out variables which can only be acquired in person such as biomarker test results. We fix our output variable to be the result of the microscopy blood smear test for malaria.

We then use Pearson's chi-squared test of independence between each of the categorical variables and the output variable (result of an individual's blood smear test) to retain the $8$ most correlated variables. The processed categorical variables are shown in the appendix.

\begin{figure}[t]
  \centering
  \includegraphics[width=0.8\linewidth]{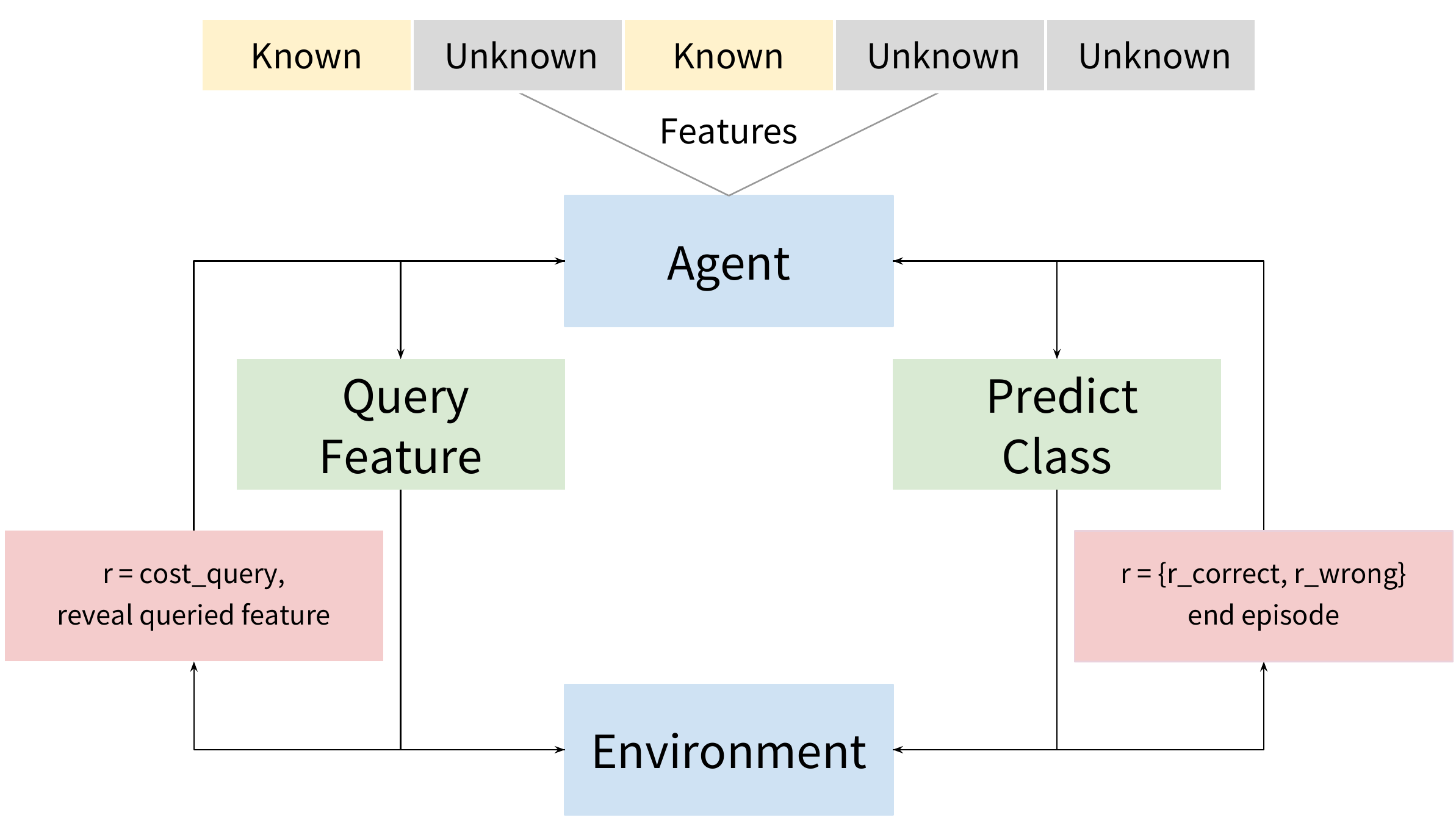}
  \caption{
   	The agent interacts with the environment, deciding whether to perform additional queries before making a prediction.
  }
  \label{fig:interaction}
\end{figure}

\section{Setup}
The agent is tasked with a $c = 2$ class prediction problem, but does not have access to any of its input features (all initially $0$). Instead, it can query any of its input features $F_n = [f_0,..,f_m], m = 8$, one at a time from the environment at a cost $cost_{query} = -0.05$. The agent querying its input feature $f_i$ is analogous to asking and getting the response to survey question $q_i$.

Within $T = k$ queries, the agent must make its prediction, and incurs a reward of $r_{correct} = +1$ if correct, and $r_{wrong} = -1$ if wrong. Making a prediction terminates the episode. If the agent makes a prediction before $2$ timesteps, or does not make a prediction before the episode ends at $t = k$, it receives the penalty of making a wrong prediction. Finally, we setup the problem such that future rewards are undiscounted ($\gamma=1$).

\section{Reinforcement Learning Agent}
We train a deep Q-Network (DQN) \cite{mnih2013playing} to learn policies directly from the results of the input feature queries made so far. The DQN outputs a $Q$ value for every action in the action space, defined such that there is an action associated with each possible query ($k$), and one with each possible prediction class ($c$), a total of $k + c$ actions. The DQN uses a neural network architecture with a $1 \times 1$ convolutional layer \cite{DBLP:journals/corr/SzegedyLJSRAEVR14} followed by $3$ fully connected layers. Each layer has $k + c$ hidden units, with a ReLU activation after each hidden layer. 

We train the DQN following the work of \cite{mnih2013playing}, using experience replay and a target network to smooth out learning and avoid divergence of the parameters. We use the Adam optimizer \cite{kingma2014adam} with minibatches of size $32$, and a learning rate that is annealed linearly from $0.00025$ to $0.00005$. The behavior policy during training is $\epsilon-$greedy with $\epsilon$ linearly annealed from $1$ to $0.01$ over the first $50k$ training samples, and fixed at $0.01$ thereafter. We train for a total of $100k$ samples, using a replay memory of the most recent $5k$ samples.

\section{Supervised Learning Baseline}
The SL baseline fixes the set of queries that it uses beforehand, and can be interpreted as a non-adaptive policy in which the survey question responses do not affect either the choice or the number of questions asked. It uses a neural network with the same architecture as the RL network, except that the final layer only has $c$ outputs: the network learns to map from the query results to the correct prediction class.

We train the network to minimize the cross-entropy loss between the predictions and the ground truth. Training examples are sampled from the training split of $~8k$ examples with equal probability assigned to each class. We use the Adam optimizer \citep{kingma2014adam} with minibatches of size $32$, with a learning rate that is annealed linearly from $0.0025$ to $0.0005$ over $20$ epochs.

\begin{table*}[t]    
\centering
\begin{tabular}{l p{15mm} p{15mm} p{15mm}}
\toprule
Model & Accuracy & Avg. Queries & Avg. Episode Reward \\
\midrule
SL network ($k = 2$) & 0.73 & 2 & +0.36 \\
SL network ($k = 4$) & 0.77 & 4 & +0.35 \\
SL network ($k = 8$) & 0.80 & 8 & +0.20 \\
RL network ($kmax = 2$) & 0.78 & 2 & +0.45 \\
RL network ($kmax = 4$) & 0.78 & 2.30 & +0.45\\
RL network ($kmax = 8$) & 0.80 & 2.35 & +0.47\\
\bottomrule
\end{tabular}
\caption{The RL networks outperform the SL networks, not only being able to make more accurate predictions, but also with fewer number of queries on average. Unlike the SL networks, which fix the set of k queries they make a priori, the RL networks are able to effectively balance the cost of making additional queries with the benefits of increased prediction accuracy.}
\end{table*}

\section{Experiments and Results}
We evaluate our models on prediction accuracy and the number of queries made. We also report the mean episode reward as defined by our setup earlier. Models are evaluated on a test set of $2k$ examples, resampled such that samples are drawn with equal probability from each class.

We investigate the effects of limiting the number of queries the agents can make. In the RL approach, we set the maximum number of queries ($kmax = {2, 4, 8}$), while in the SL approach we fix the number of queries ($k = {2, 4, 8}$). Queries are ordered by decreasing correlation with the output variable, so using $k$ queries refers to using the top $k$ correlated variables.

As $k$ increases, the SL baselines reach higher accuracies, but at the cost of an increase in the number of queries. The tradeoff at $k = 2$ and $k = 4$ produces comparable average episode scores, but the cost of making even more queries at $k = 8$ outweighs the benefits of increased accuracy.

As $kmax$ increases, the RL agents also reach higher accuracies; they are, however, able to keep the average number of queries small. The agents are therefore able to reach higher episode rewards.

\section{Related Work}
The idea of training RL agents to query before making predictions has recently been explored in the reinforcement learning literature. \citet{learningtostop} train an agent to dynamically decide whether to continue or to terminate the inference process in machine comprehension tasks. \citet{rightquestions} train an agent to systematically perturb a question to aggregate evidence until it can confidently answer it. \citet{featureselection} propose an algorithm to find which features are necessary solve a particular task.

\section{Conclusion}
We build a deep reinforcement learning (RL) agent that can predict the likelihood of an individual testing positive for malaria by asking questions about their household. We propose an algorithm to adaptively determine which survey question to ask next and when to stop to make a prediction about their likelihood of malaria based on their responses. We show that we can make predictions with high accuracy, and that adaptive surveys can significantly reduce the number of questions needed to attain the high accuracy, enabling large-scale and cost-effective monitoring of malaria through SMS surveys.

\bibliography{bib}
\bibliographystyle{bst}

\clearpage
\appendix
\section{Appendix}

\begin{table*}[ht] 
\centering
\begin{tabular}{c p{90mm} c}
\toprule
Question & Variable Name & Num Categories \\
\midrule
1 & Region & 8 \\
2 & Main floor material & 9 \\
3 & Owns land usable for agriculture & 2 \\
4 & Has electricity & 2 \\
5 & Has television & 2 \\
6 & Whether nets are used alternatively in this community & 2 \\
7 & Most people in the community sleep under an ITN all the time & 4 \\
8 & Type of place of residence & 2 \\
\bottomrule
\end{tabular}
\caption{The list of processed categorical variables, sorted by decreasing correlation with the output variable (the result of the blood smear test).}
\label{tab:processed_variables}
\end{table*}

\vspace{20mm}

\begin{figure}[ht] 
  \centering
  \includegraphics[width=0.8\linewidth]{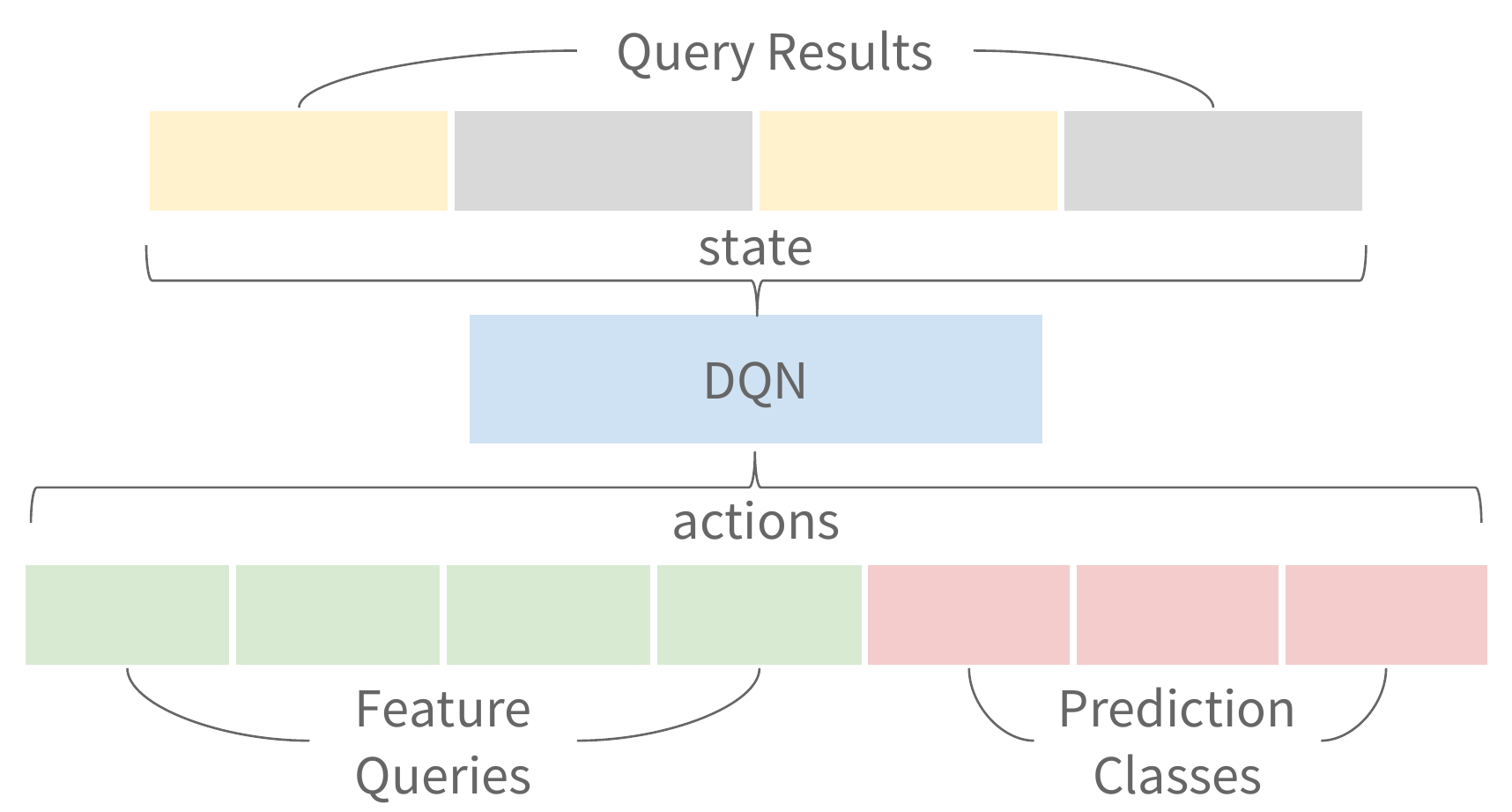}
  \caption{
  The DQN outputs a Q value for each of the possible queries and prediction outputs from its input state.
  }
  \label{fig:predq}
\end{figure}

\begin{figure}[ht] 
  \centering
  \includegraphics[width=\linewidth]{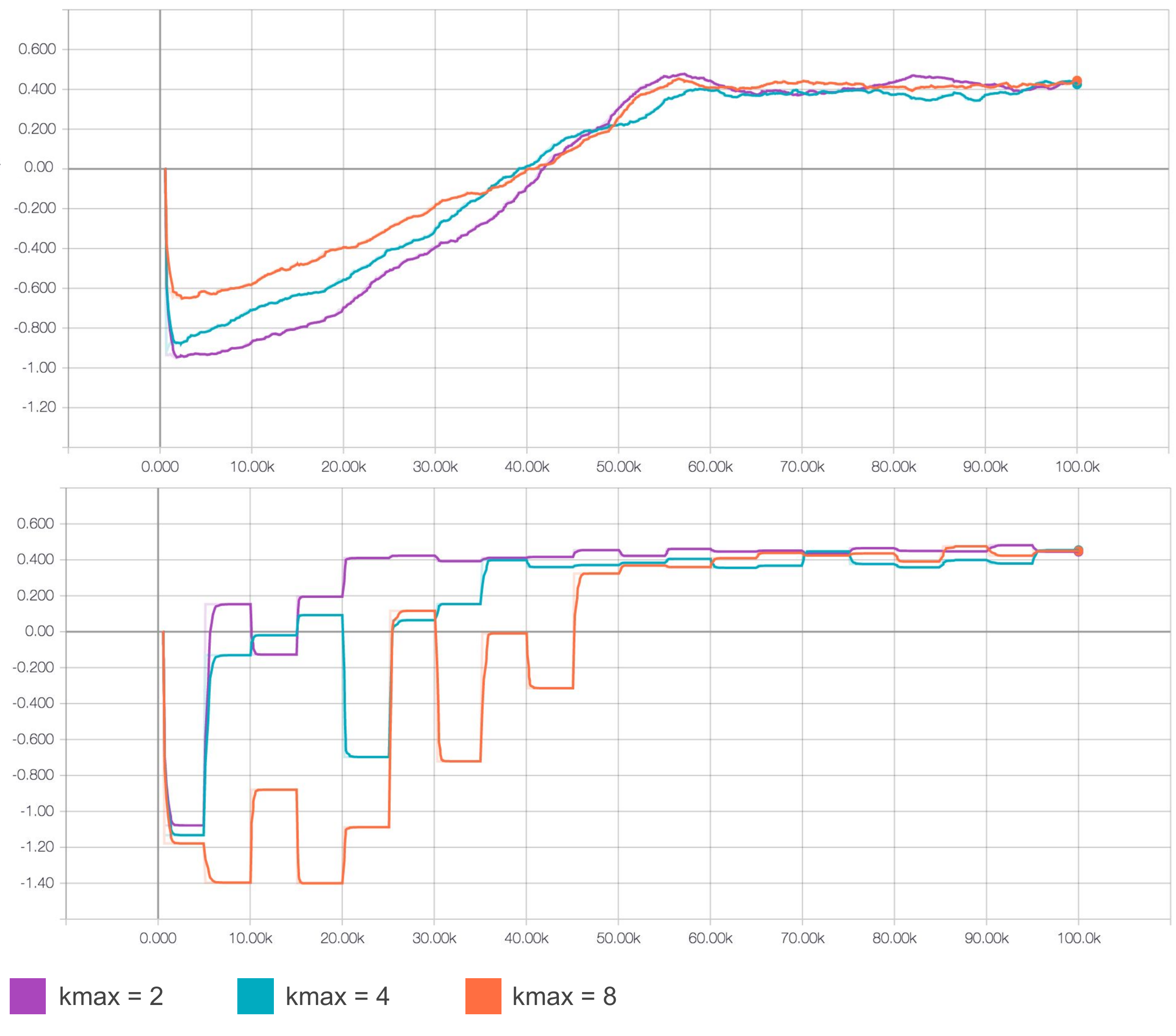}
  \caption{
   	The average training (top) and evaluation (bottom) episode rewards of the RL agents during the training process. The behavior policy for training episodes is $\epsilon-$greedy with a linearly annealed $\epsilon$; for test episodes, we set $\epsilon = 0$, and act greedily with respect to the $Q$ values outputted by the DQN.
  }
  \label{fig:train_test_reward}
\end{figure}

\end{document}